\documentclass{acmsiggraph}                     

\usepackage[scaled=.92]{helvet}
\usepackage{times}

\usepackage{parskip}

\usepackage[labelfont=bf,textfont=it]{caption}

\usepackage{soul}
\usepackage{fancyvrb}
\usepackage{times}
\usepackage{amsmath}
\usepackage{amssymb}
\usepackage{graphicx} 
\usepackage{epstopdf}

\usepackage{hyperref}
\DefineVerbatimEnvironment{code}{Verbatim}{fontsize=\small}
\DefineVerbatimEnvironment{example}{Verbatim}{fontsize=\small}

\onlineid{0}

\title{A Novel Method for Vectorization}

\author{Tolga Birdal\thanks{e-mail: tbirdal@gravi.com.tr}\\ Gravi Ltd. %
\and Emrah Bala\thanks{e-mail: emrah@gravi.com.tr}\\ Gravi Ltd. %
}


\begin{document}

\teaser{
$
\begin{array}{cc}
\includegraphics[width=3in]{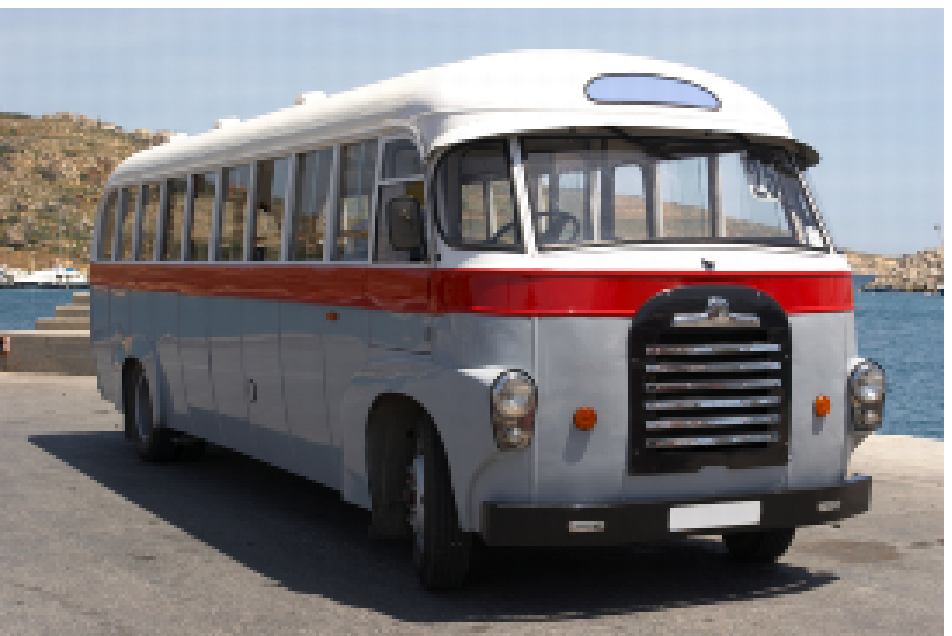}&
\includegraphics[width=3in]{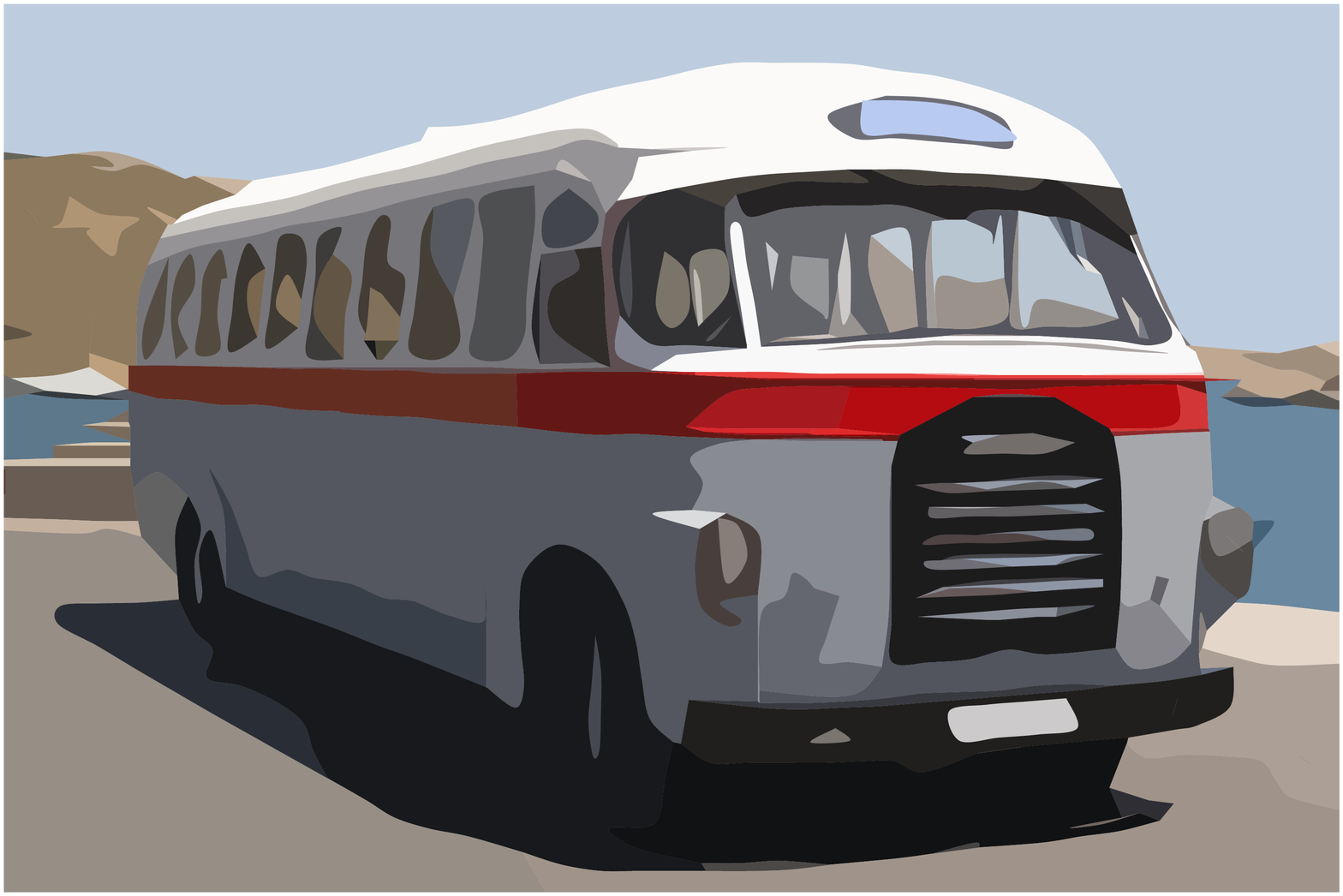}\\
\mbox{a}&\mbox{b}\\
\end{array}$
\caption{Vectorization in a customizable and aesthetical way: (a) Input raster image, (b) Sample vector output obtained using segmentation and Catmull Rom spline approximation}
\label{fig:teas}
}


\maketitle


\begin{abstract}

Vectorization of images is a key concern uniting computer graphics and computer vision communities. In this paper we are presenting a novel idea for efficient, customizable vectorization of raster images, based on Catmull Rom spline fitting. The algorithm maintains a good balance between photo-realism and photo abstraction, and hence is applicable to applications with artistic concerns or applications where less information loss is crucial. The resulting algorithm is fast, parallelizable and can satisfy general soft realtime requirements. Moreover, the smoothness of the vectorized images aesthetically outperforms outputs of many polygon-based methods.

\end{abstract}


\keywordlist

\section{Introduction}
By the advancements in digital photography, image processing and computer vision techniques, many file formats for representing raster images become available. Indeed, most of these image formats are no different than each other, since they all use pixels representations. However, representing images in pixels brings certain drawbacks. Being non-scalable is the main problem that the entire digital art world complains about. However, there are more disadvantages then there seems to be. For example, people working in the field of digital art may desire a more non-realistic image representation which is smoother or more cartoon-like with certain artistic modalities. In addition, they may require more control over different structures or segments in the image. Digital artists have more tendency in using curved structures other than polygons. That would give them more control over the drawing and a rather smooth look, which cannot be achieved by polygonal rendering techniques. Apart from these possible artistic expectations, one may require a more mathematical representation of what he actually sees and perceives in a photograph. In all of these cases, vector representations of images can come handy.

Vector graphics use geometrical primitives to express a raster image. With the help of these primitives, they happen to be more compact, editable, scalable, resolution-independent and even smaller in file size. Such properties also make them amenable for portable applications. Vectorization, which is the name given to the technology to convert raster images into vector formats, paves the way for expanding the area of usability of vector images, and their integration to digital art.

This paper proposes a novel raster to vector conversion algorithm which is highly capable, efficient, extensible and customizable. It doesn't solve any linear systems, doesn't perform any non-linear optimizations and it doesn't use complicated data structures like meshes. For the reasons above, the algorithm has a relatively low computational cost, and it is easy to implement. Because the method takes advantage of Catmull Rom splines, the resulting images have a smooth look, which is desirable in most of the cases. Certain additional attempts have been made to reduce the final size of the generated vector file. Although, generated vector images are not restricted to be exported in a specific vector file format, we used SVG (Scalable Vector Graphics) in this paper. Those images can also be exported in many other common formats such as EPS (Encapsulated PostScrit) or PDF (Portable Document Format).

Our algorithm relies on a relatively good segmentation method, where many methods can be used depending on the purpose of the vectorization. By using over-segmented and under-segmented images, the user can preserve more details for realistic styles or consciously lose information for cartoon-like results, respectively. After segmenting the input image, the rest of the algorithm concentrates on 3 main problems: Finding the segment boundaries, smoothing the segment boundaries by splines and reformatting these splines for generation of vector files. Finally, an efficient, parallelizable algorithm for generating smooth vectors is achieved.

This paper is organized as follows: section 2 reviews the relationship of this work to previous approaches, section 3 explains our method step by step, section 4 offers sample segmentation methods for different look and feel, section 5 provides vectorization results and evaluates them. Finally we conclude in section 6 with a summary of the algorithm presented and a discussion of the future work.

\section{Background}
The history of vectorization process begins from scanned-line art, black-white images and continues to evolve to maps, cartoon drawings and finally realistic images. Various approaches have been proposed to solve the same problem. Unfortunately, most of these works focus on line drawings, fonts, texts, and sketches which are in fact binary images with a sparse set of pixels in the region of interest \cite{potrace}, \cite{Fan95}, \cite{VLD}. Others use simpler algorithms that try to use only polygons, or polylines to describe the vector drawing \cite{svgstat}. Even though there are improvements made in terms of output sizes, adaptive vertex selection, these methods lack the smoothness and have little artistic aspects \cite{svgstat}, \cite{Watershed}. There are more sophisticated algorithms used in tools like Adobe Live Trace \cite{adobe09}. However, there are 2 main downsides in using such methods. First of all, even though they especially target the digital art community, the algorithms proposed by Adobe or CorelDraw require too much parameter tuning. There even exist detailed tutorials on how to obtain good results. Second downside is that these algorithms are dependent on a single, integrated segmentation algorithm that's not very customizable. It also limits the creativity and customizability. There are certain other methods proposed to preserve the gradient information in segmentation, such as topology preserving gradient meshes. However, the improvements in that technology require non-standard vector drawing structures and definitions, which are obviously not supported in any of SVG, EPS or PDF. Because generally non-linear optimizations are of concern the efficiency of these algorithms are low, in general \cite{gradientmesh}.

\section{Our Approach}

Our approach is basically based on splitting the image into regions by using an image segmentation method, and representing these regions by curve sets each of which represents a closed boundary. 

Image segmentation methods used for this work are briefly discussed in Chapter 4. The order of segmentation determines the level of detail in vector output, i.e.\ an over-segmented image produces a high detail vector output, whereas an undersegmented image produces a low detail vector output. 

Suggested vectorization method is composed of four main operations applied on segmentation output, sequentially. First of all, boundaries between regions are generated; secondly, the boundary pieces around each region is determined; thirdly, some points on these pieces are sampled and splines are fitted on these points; and finally, these curves are sorted and written in an SVG file as closed curves with specific fill colors. These steps are explained in detail in following subsections.

\subsection{Generating Subpixel Region Boundaries}
\begin{figure}[h]
\begin{center}
  \includegraphics[width=0.4\linewidth]{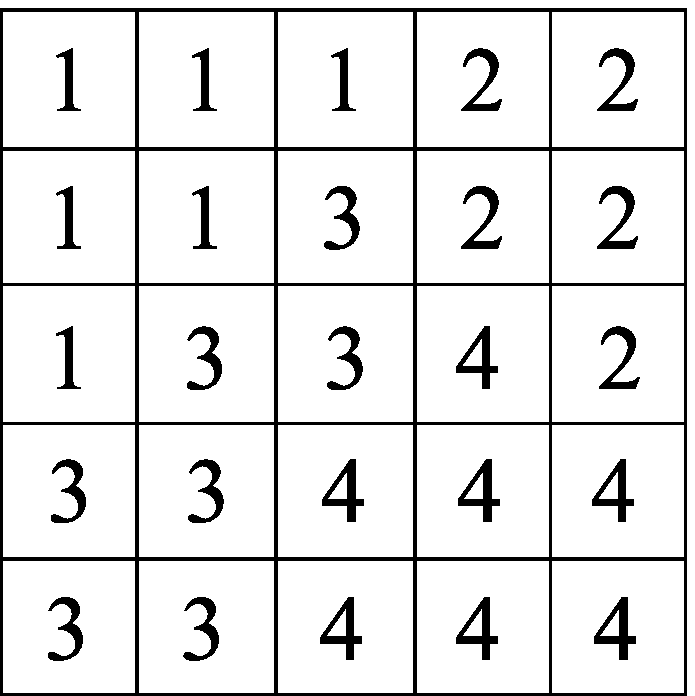}
\end{center}
   \caption{A sample segmentation output}
\label{fig:one}
\end{figure}

A segmentation output {$I$} is nothing but a label image which stores the label ($i$) of corresponding region ({$R_i$}) for each pixel with coordinates {$(x,y)$}, where {\em $x=0,1,..,w_I-1$} and {$y=0,1,..,h_I-1$}. {$w_I$} and {$h_I$} are width and height of {$I$}, respectively. Output of a sample segmentation, with {$w_I=5$} and {$h_I=5$}, is shown in Figure~\ref{fig:one}.

A subpixel boundary image {$S$} is initialized with width ($w_S$) and height ({$h_S$}) such as: {$w_S=2 \times w_I-1$} and {$h_S=2\times h_I-1$}. {$S$} is a binary image which shows nothing but subpixel edges between pixels with different label values in {$I$}, and it is defined as follows:
\begin{equation}
S(x,y) = \left\{ \begin{array}{rc}
   &\mbox{(if (x is odd and y is even)}\\
	&\mbox{and}\\
	&\mbox{if ($I((x+1)/2,y/2)\neq$}\\ 
	&\mbox{$I((x-1)/2,y/2)))$} \\
	1 &\mbox{or} \\
	&\mbox{(if (x is even and y is odd)}\\
	&\mbox{and}\\
	&\mbox{ if ($I(x/2,(y+1)/2)\neq$}\\ 
	&\mbox{$I(x/2,(y-1)/2)))$} \\
\\
  0 &\mbox{ otherwise}
\end{array}  \right.
\label{eq:eq1}
\end{equation}
where $x=0,1,..,w_S-1$ and $y=0,1,..,h_S-1$. 

\begin{figure}[h]
\begin{center}
\includegraphics[width=0.4\linewidth]{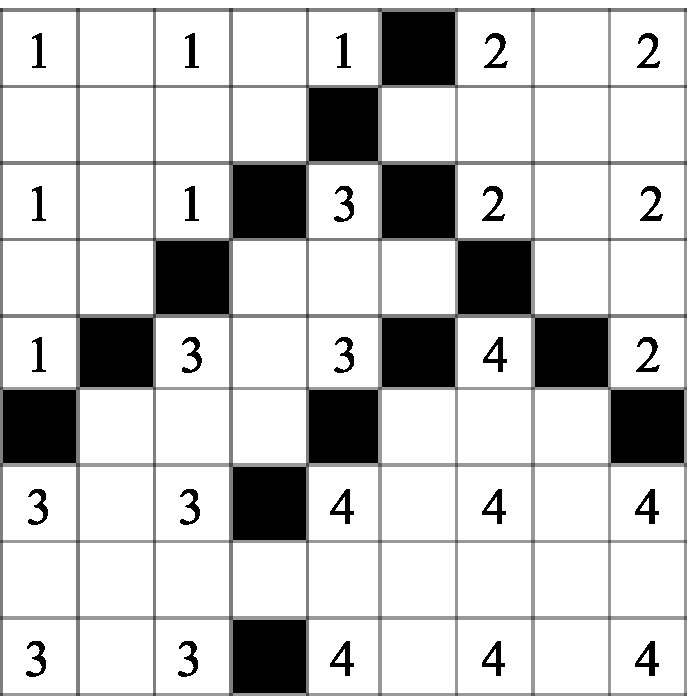}
\end{center}
   \caption{A sample subpixel edge image.}
\label{fig:two}
\end{figure}

By using (\ref{eq:eq1}), the found edge pixels of sample segmentation output shown in Figure~\ref{fig:one} is given in Figure~\ref{fig:two}. The labels of regions are also shown on the subpixel edge image in Figure~\ref{fig:two}.

Furthermore, $S$ is processed to fill the gaps between edge pixels and find {\emph{junction points}}, which are of high importance for the remaning procedure of our method. The gaps are filled to obtain 8-connected boundary pieces (\ref{eq:eq2}).
\begin{equation}
S(x,y) = \left\{ \begin{array}{cc}
   &\mbox{ if ($S(x+1,y)=1$ and} \\
	&\mbox{$S(x-1,y)=1$ }) \\
	1 &\mbox{or} \\
	 &\mbox{ if ($S(x,y+1)=1$ and}\\ 
&\mbox{$S(x,y-1)=1$}) \\
\\
  S(x,y) &\mbox{ otherwise}
\end{array}  \right.
\label{eq:eq2}
\end{equation}
where $x=1,3,..,w_S-2$ and $y=1,3,..,h_S-2$, i.e.\ (\ref{eq:eq2}) is evaluated for only subpixel locations which are between potential edge pixels found by (\ref{eq:eq1}).

A \emph{junction point} is a point where more than two regions are neigbour to each other. This is valid for locations having more two neighbour edge pixels. Edge pixels are classified depending on this condition as shown in (\ref{eq:eq3}).
\begin{equation}
J(x,y) = \left\{ \begin{array}{rc}
1   &\mbox{if $S(x+1,y)+S(x-1,y)+$} \\
	&\mbox{   $S(x,y+1)+S(x,y-1)>2$} \\
\\
  0 &\mbox{ otherwise}
\end{array}  \right.
\label{eq:eq3}
\end{equation}
where $x=1,3,..,w_S-2$ and $y=1,3,..,h_S-2$.

\begin{figure}[h]
\begin{center}
\includegraphics[width=0.4\linewidth]{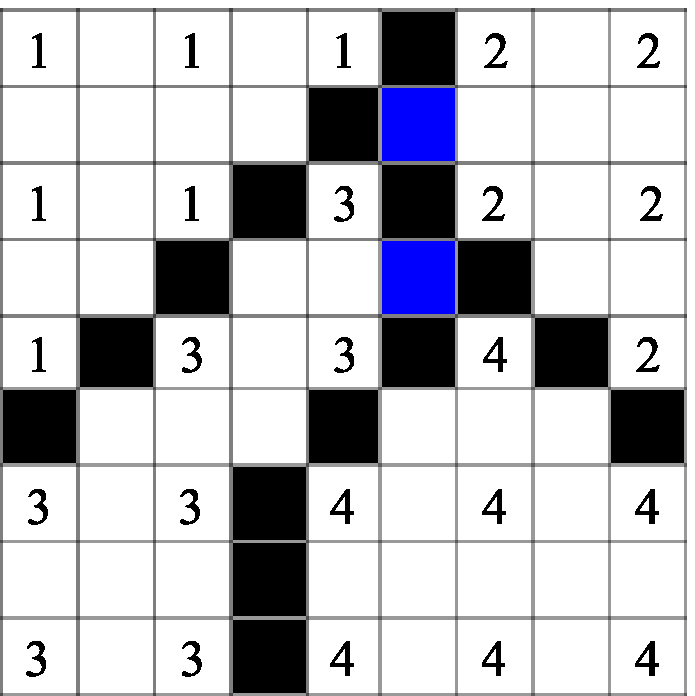}
\end{center}
   \caption{A sample final subpixel edge image.}
\label{fig:three}
\end{figure}

A sample final subpixel edge image is shown in Figure~\ref{fig:three}, on which the gaps are filled according to (\ref{eq:eq2}) and junction points are marked in blue according to (\ref{eq:eq3}).

\subsection{Extracting Boundary Pieces}
A boundary piece $b_k$, which is nothing but a connected and ordered list of points, is a curve segment virtually seperating two different regions $R_i$ and $R_j$ in $S$. A boundary can exist in three different forms: as a curve between two junction points, as a single closed curve, as a curve starting and ending in borders of $S$.

Firstly, $S$ is processed to extract the boundary pieces between junction points found by using (\ref{eq:eq3}). Then, $S$ is searched for other two possible forms of pieces. Extraction step includes storing the list of points in every piece by using a classical 8-connected curve tracing algorithm. Closed pieces are stored in a different list, since they will be rendered at last for them not to be over-painted by an encapsulating region, in final vector image.

In addition, a list $N_i$ for each region $R_i$ in $S$ is generated. While extracting each {\emph {open}} boundary piece $b_k$, its index $k$ is pushed into the lists $N_i$ and $N_j$ of separated regions, which are $R_i$ and $R_j$, respectively. It is done to store the information of enclosing pieces for every region. 
\subsection{Fitting Spline On A Boundary Piece}

A spline is a piecewise polynomial function used to approximate a shape by fitting curve on an ordered set of points. It is widely used in computer graphics, since they are easy to construct and evaluate. In addition, it is an accurate way of representing more complex shapes by fitting smooth curves. That is why, fitting a spline on a predetermined list of points is a feasible solution to approximate a boundary piece in a smooth and efficient way.  

A spline curve is evaluated by using the parametric equation given in (\ref{eq:eq5}).
\begin{equation}
\begin{array}{l}
C(u)=UMP^T\\
\end{array}
\label{eq:eq5}
\end{equation}
where $U$ is the parameter vector defined as $U=\left[\begin{array}{llll}u^3 & u^2 & u & 1\end{array}\right]$, $P$ is the control point vector on which the spline curve will be fitted and it is defined as $P=\left[\begin{array}{cccc}p_{-1} & p_0 & p_1 & p_2\end{array}\right]$, and $M$ is the characteristic matrix of used spline.

A Catmull-Rom Spline is a local cubic spline and it can be specified with the matrix $M_{catmull}$ given in (\ref{eq:eq4}). Catmull-Rom Spline has C1 continuity, local control, and interpolation, but does not lie within the convex hull of their control points. 
\begin{equation}
M_{catmull}=\frac{1}{2}\left[ \begin{array}{rrrr}
  -1 & 3 & -3 & 1\\
  2 & -5 & 4 & -1\\
  -1 & 0 & 1 & 0\\
  0 & 1 & 0 & 0 
  \end{array}\right]
\label{eq:eq4}
\end{equation}

For Catmull-Rom situation the spline curve is calculated using (\ref{eq:eq5}) from $p_0$ to $p_1$ by incrementing $u$ from 0 to 1.

Catmull-Rom Spline passes from its control points \{$p_0$,$p_1$\}, which is not a common property for all types of splines. This property makes Catmull-Rom Spline appropriate for our situation since fitted curve has to pass from the junction points in order not to leave gaps between regions. This situation will be explained in detail in section 3.4.


To represent a boundary piece by a spline, curve points are sampled uniformly. The end points of the boundary piece should be selected as control points independent of the sampling resolution.

In vector images, generally, Cubic Bezier Pieces are used to represent curves. A Cubic Bezier Piece is defined by four control points: $P_b=\left[\begin{array}{cccc}p^b_{-1} & p^b_0 & p^b_1 & p^b_2\end{array}\right]$ and it passes only from $p^b_{-1}$ and $p^b_2$. $p^b_0$ and $p^b_1$ are used to define the shape of the curve. At this point, we should find a method to represent a Catmull-Rom Spline curve by Cubic Bezier Pieces.

A Cubic Bezier Piece can be specified with the matrix $M_{bezier}$ given in (\ref{eq:eq6}). Cubic Bezier Piece is also evaluated parametrically by using (\ref{eq:eq5}).
\begin{equation}
M_{bezier}=\left[ \begin{array}{rrrr}
  -1 & 3 & -3 & 1\\
  3 & -6 & 3 & 0\\
  -3 & 3 & 0 & 0\\
  1 & 0 & 0 & 0 
  \end{array}\right]
\label{eq:eq6}
\end{equation}

To represent a Catmull-Rom Spline curve segment ($C_c$) between $p^c_0$ and $p^c_1$ and defined by control points $\left[\begin{array}{cccc}p^c_{-1} & p^c_0 & p^c_1 & p^c_2\end{array}\right]$, a Cubic Bezier Piece ($C_b$) defined by control points $\left[\begin{array}{cccc}p^b_{-1} & p^b_0 & p^b_1 & p^b_2\end{array}\right]$ should be calculated. It is clear that, $p^b_{-1}=p^c_0$ and $p^b_{2}=p^c_1$ to fix the end points of both curves. Two extra control points $p^b_{0}$ and $p^b_{1}$ should be generated for $C_b$.

The parametric equations for $C_c$ and $C_b$ are equal to each other as shown in (\ref{eq:eq7}), since they define the same curve. $P_b$ is found as (\ref{eq:eq8}) by doing necessary cancellations on (\ref{eq:eq7}).
\begin{eqnarray}
 C_b &=& C_c     \nonumber  \\
  UM_{bezier}P_b^T &=& UM_{catmull}P_c^T \label{eq:eq7} \\
  M_{bezier}P_b^T &=& M_{catmull}P_c^T \nonumber \\
  P_b^T &=& M_{bezier}^{-1}M_{catmull}P_c^T \nonumber\\
  P_b &=& (M_{bezier}^{-1}M_{catmull}P_c^T)^T \label{eq:eq8}
\end{eqnarray}
(\ref{eq:eq8}) is evaluated for only the elements $p^b_0$, $p^b_1$ of $P_b$. Therefore, (\ref{eq:eq8}) can be decomposed into two linear equations given in (\ref{eq:eq9}) and (\ref{eq:eq10}).
\begin{equation}
p^b_0=-\frac{p^c_{-1}}{6}+p^c_0+\frac{p^c_1}{6}
\label{eq:eq9}
\end{equation}
\begin{equation}
p^b_1=\frac{p^c_0}{6}+p^c_1-\frac{p^c_2}{6}
\label{eq:eq10}
\end{equation}

Furthermore, C1 continuity (continuity of first derivative) should be satisfied at common points of consecutive pieces, to obtain a smooth curve. This is achieved by keeping the distance $D_{-1,0}$ between $p^b_{-1}$ and $p^b_0$ of a piece equal to the distance $D_{1,2}$ between $p^b_1$ and $p^b_2$ of the previous piece, where $p^b_{-1}$ of the current piece and $p^b_2$ of the previous piece are actually the same point. Therefore, after calculating $p^b_0$ and $p^b_1$ for first Cubic Bezier Piece, $p^b_0$ of following pieces are calcuated using the C1 continuity constraint (\ref{eq:eq11}) instead of using (\ref{eq:eq9}).
\begin{equation}
p^b_0=2\times p^b_{-1}-p^{'b}_1
\label{eq:eq11}
\end{equation}
where $p^{'b}_1$ is from the previous piece.

The above methodology can be used to convert different types of splines by using their $M$ matrix. For our situation, it is enough to convert Catmull-Rom Spline to a Cubic Bezier Piece.

\subsection{Writing SVG Vector Image}
An SVG image is basically formed by using primitive types of geometrical shapes, their stroke width values, stroke color values and fill color values. For our method, closed boundaries of regions will be represented by paths composed of several curves and lines. A line between ($x1,y1$) and ($x2,y2$) is added to a path by code segment:
\begin{code}
x1,y1 L x2,y2
\end{code}
whereas, a Cubic Bezier Piece is added in the following way:
\begin{code}
x1,y1 C x2,y2 x3,y3 x4,y4
\end{code}
where $P_b=\left[\begin{array}{cccc}(x1,y1) & (x2,y2) & (x3,y3) & (x4,y4)\end{array}\right]$.


Up to this point, boundary pieces around each region is determined and stored in a list $N_i$ for each region $R_i$. In addition, for regions, which are neighbour to the image borders, image border piece around each region $R_i$ is determined and pushed to its neighbour list $N_i$.

For each region, the open boundary pieces in the list $N_i$ are sorted according to their start and end points to form a closed region boundary together. After that, control points, which are necessary to fit curves, are uniformly sampled (start and end points of the pieces are preserved by explicitly inserting them as control points). Uniform sampling is a process which actually randomizes the shape of the boundary. For different purposes, different sampling strategies can be evolved. 

$x$ and $y$ coordinates of sampled points are multiplied by $0.5$ since they are obtained from subpixel edge image. For boundary pieces represented by only two control points lines are used instead of Cubic Bezier Pieces, whereas for boundary pieces represented by more than two control points, Catmull-Rom to Cubic Bezier Piece conversion, which is described in section 3.3, is performed and Cubic Bezier Pieces are defined with calculated points.

All curves and lines forming a closed boundary are concatenated as a single \emph {path} statement in the SVG file. The fill color is automatically selected depending on the enclosed region (mean color, median color, etc.).

Furthermore, the \emph {closed} boundary pieces, which were stored in a seperate boundary piece list as told in section 3.2, are actually holes in previously drawn regions. Therefore, they are written in SVG file at last in order them to be visible in the vector image.

\section{Image Segmentation}
This paper presents 3 choices of segmentation methods. These methods are chosen because they are fast, and produce artistically varying results. 


\subsection{Statistical Region Merging (SRM)}
In SRM \cite{SRM04}, image is considered to be a perfect unknown scene. In such a scene pixels are represented by a set of distributions from which the colors are sampled. This formulation can also be referred as an inference problem, where the statistical regions are seeked.  The resulting regions are assumed to obey the following homogeneity constraints:
	 a) Inside any statistical region and given any color channel {R,G,B} the statistical pixels have the same expectation for this color channel.
	 b) The expectations of adjacent statistical regions are different for at least one color channel ${R,G,B}$.

After defining such rules, using the independent bounded difference inequality Nock and Nielsen derived the following similarity predicate:

\begin{equation}
P(R,R')=
\begin{cases}
true, &  \mbox{if } |\overline{R'_i}-\overline{R_i}| \leq \sqrt{b^2(R) + b^2(R')} \\
  false, & \mbox{otherwise} \\
\end{cases}
\end{equation}

where
\begin{equation}
b(R)=\sqrt{ \dfrac{1}{2Q|R|} ln(\dfrac{|\mathcal{R}_{|R|}|}{\delta})}
\end{equation}

For the sorting of pixels before merging the similar regions, Nock and Nielsen propose a maximization of a distance function:

\begin{equation}
f (p, p')= max_{a \in R,G,B}\ f_a (p, p') \\
\end{equation}

For the choice of $f_a$ a direct absolute difference method is applied (Also used in our experiments).  Other than that, a sobel gradient of the image can also be used. Here is a quick overview of our implementation:
1) 	In the beginning, each pixel forms one region
2) 	Setup the observed mean values for each channel: R=I at the beginning
3) 	Obtain edge pairs by traversing the image and using south and east neighbors, and record difference
4) 	Sort the edges according to $f(p,p')$
5) 	Merge the similar regions  (union-find) while updating observed means according to equation.
6) 	Merge the small regions  (union-find) while updating observed means
7) 	Obtain the label map

\subsection{Color Structure Coding (CSC)}

CSC \cite{CSC03} is a hierarchical region-growing technique performed on a special hexagonal topology, constructed by islands of different levels. It consists of two phases which are linking phase (Hexagonal Topology representation) and segmentation phases. Below we shortly explain these phases.

\subsubsection{Hexagonal Topology}
\begin{figure}[h]
\begin{center}
 \includegraphics[width=0.5\linewidth]{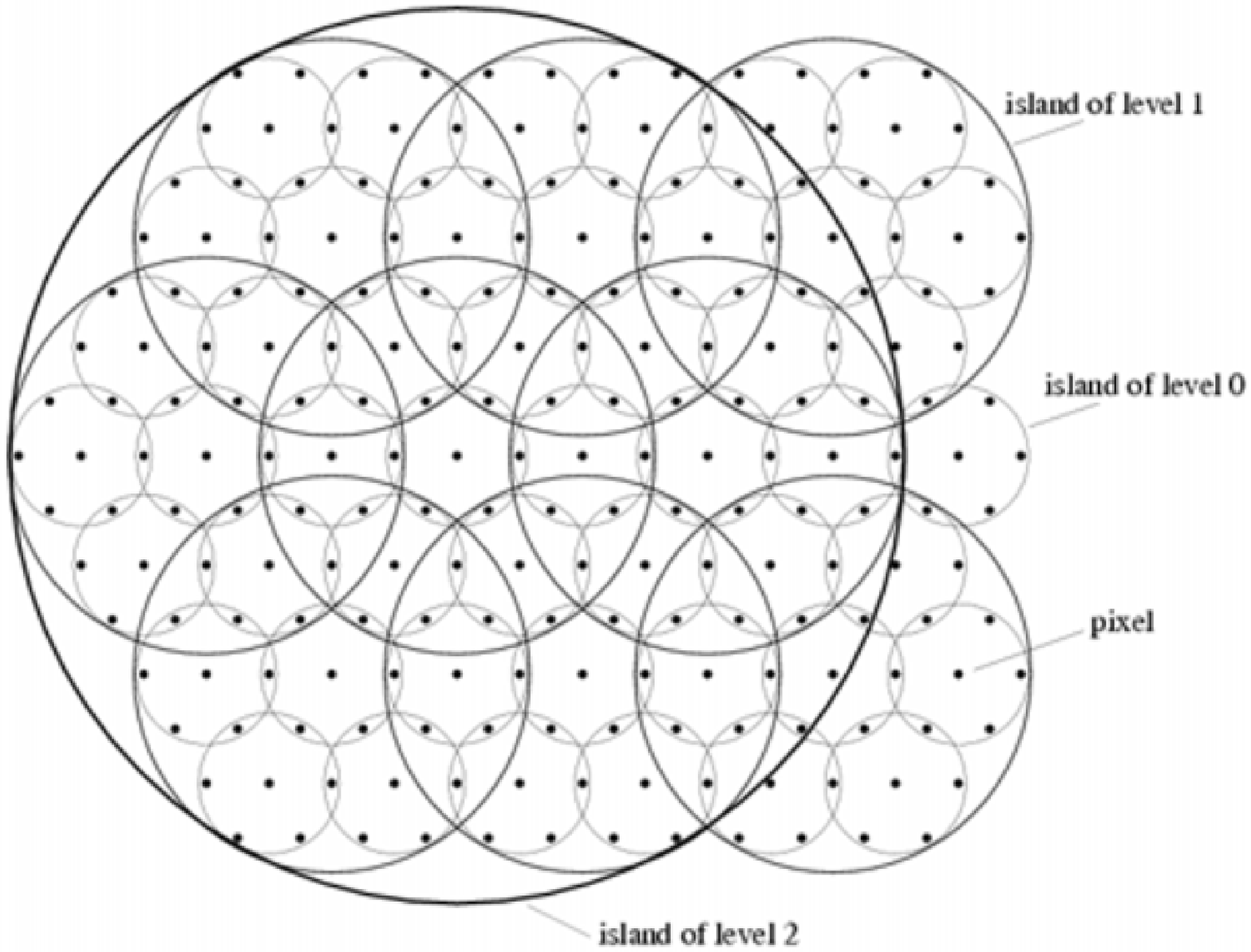}
\end{center}
   \caption{Hexagonal Topology. Only Island Level 3 is shown.}
\label{fig:five}
\end{figure}

In Figure~\ref{fig:five}, the hexagonal topology of the islands is presented. the smallest circles denote the pixels and each 7 pixels are grouped together to form 0th Island Level. Seven 0 Level Islands are composed to form 1st Island Level. This procedure is repeated until the entire image is grouped into Island Level N, where N is the last island level. This iterative algorithm is called the linking phase.

Applying this structure to an orthogonal bitmap could be tricky. Because this topology cannot represent the orthagonality, y axis of the hexagonal lattice is scaled and each second row is shifted by half unit to left. Therefore, the shapes of the islands are altered. This new representation is consistent with the hexagonal topology and the distorted islands can directly be used in orthogonal lattice. 
The intersection of a level N island with another level N island always includes a level N-1 island. The smallest island is referred as a pixel. 

\subsubsection{Segmentation}
In the segmentation phase the pixels in similar colors in an island are grouped into code elements. The similarity measure is simply defined by a binary threshold mapping. If Euclidean distance between two colors is smaller than a certain threshold, the images are said to be similar. Increasing this value leads to under-segmentation. Other color metrics could also be defined. This mapping is iterated through the entire hierarchy. 

Because CSC is a hierarchical representation CSC is really fast and successful in preserving global features. These properties make it one of the best algorithms in terms of speed and robustness.

\subsection{Graphs}

A very straightforward approach which is similar to SRM is simple graph based segmentation \cite{GBS}. Being also efficient, this technique produces more abstract results than SRM and CSC. It also uses a pair-wise comparison predicate (D) , which is defined as

\begin{equation}
D(C_1,C_2)= 
 \begin{cases}
  true, &  \mbox{if } D_i f(C_1, C_2)>MInt(C_1, C_2) \leq 0 \\
  false, & \mbox{otherwise} \\
 \end{cases}
\end{equation}

Where C's being different components (regions) and the minimum internal difference (Mint) is defined as

\begin{equation}
MInt(C_1, C_2)=min(Int(C_1)+\tau(C_1),Int(C_2)+\tau(C_2)) \\
\end{equation}
\begin{equation}
D_i f(C_1, C_2)=min_{v_i \in C_1, v_j \in C_2, (v_i, v_j) \in E}\ w(v_i, v_j)
\end{equation}

And internal difference:
\begin{equation}
Int(C)=max_{e \in MST(C,E)}\ w(e)
\end{equation}

where w denote the edge weights and MST the minimum spanning tree of the graph. The threshold function $\tau(T)$ exists to compensate for small components and thus is defined as $\tau(C)=k/|C|$ with k being a constant and $|C|$ the size of C. Please note that, smaller components are allowed when there is a sufficiently large difference between neighboring components.

The common ground which's shared by all these segmentation methods is that they all output for each region the list of pixels belonging to it. The rest of the algorithm uses this output as its input.

\section{Results}

\subsection{Quality Evaluation}

Output quality of suggested vectorization algorithm is evaluated in two different aspects. Since we mainly consider the artistic aspect of vectorization operation, evaluating the output vector images in aesthetics manner is primary corcern of this work. 

\begin{figure}[h]
\begin{center}
$
\begin{array}{cc}
\includegraphics[width=0.4\linewidth]{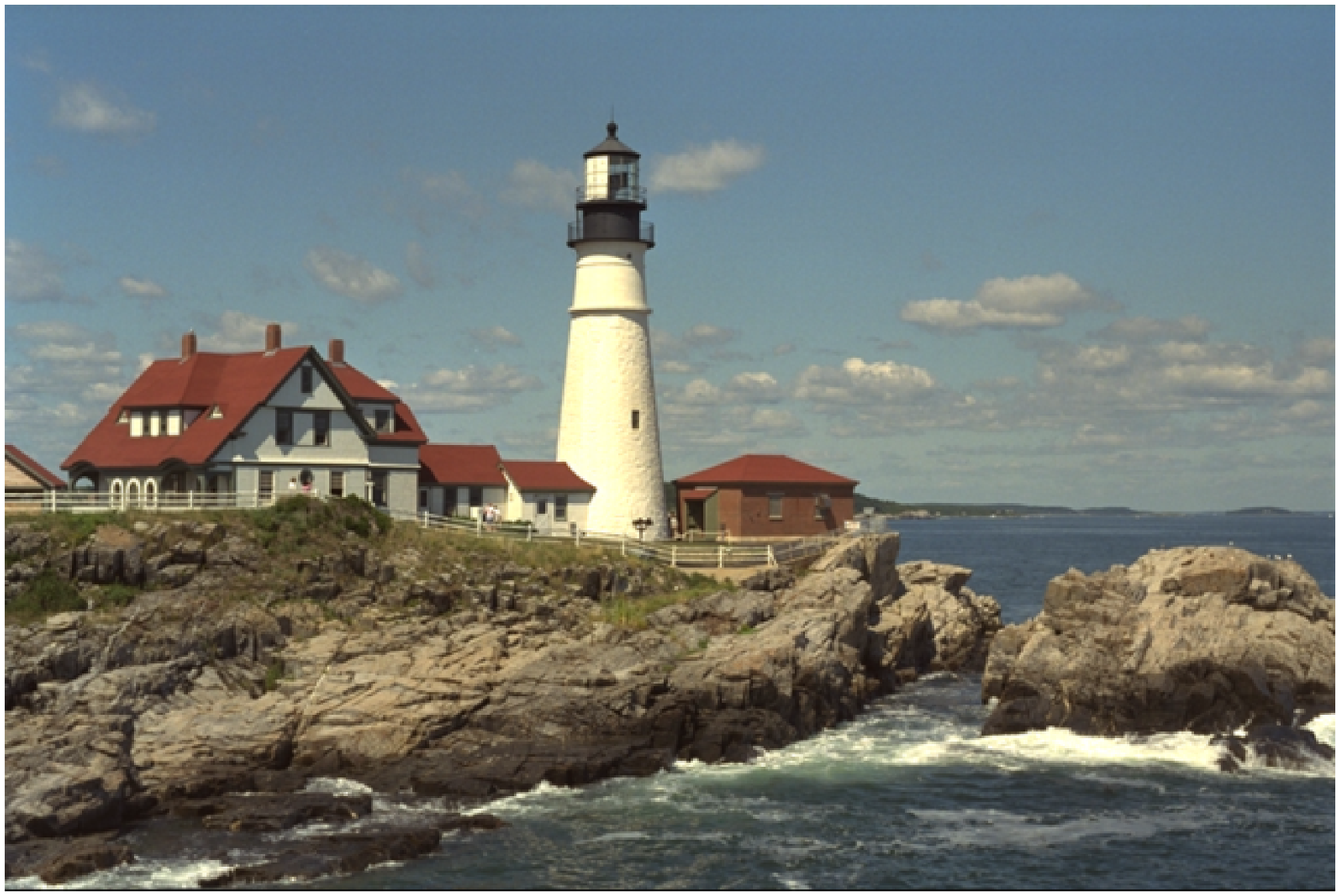}&
\includegraphics[width=0.4\linewidth]{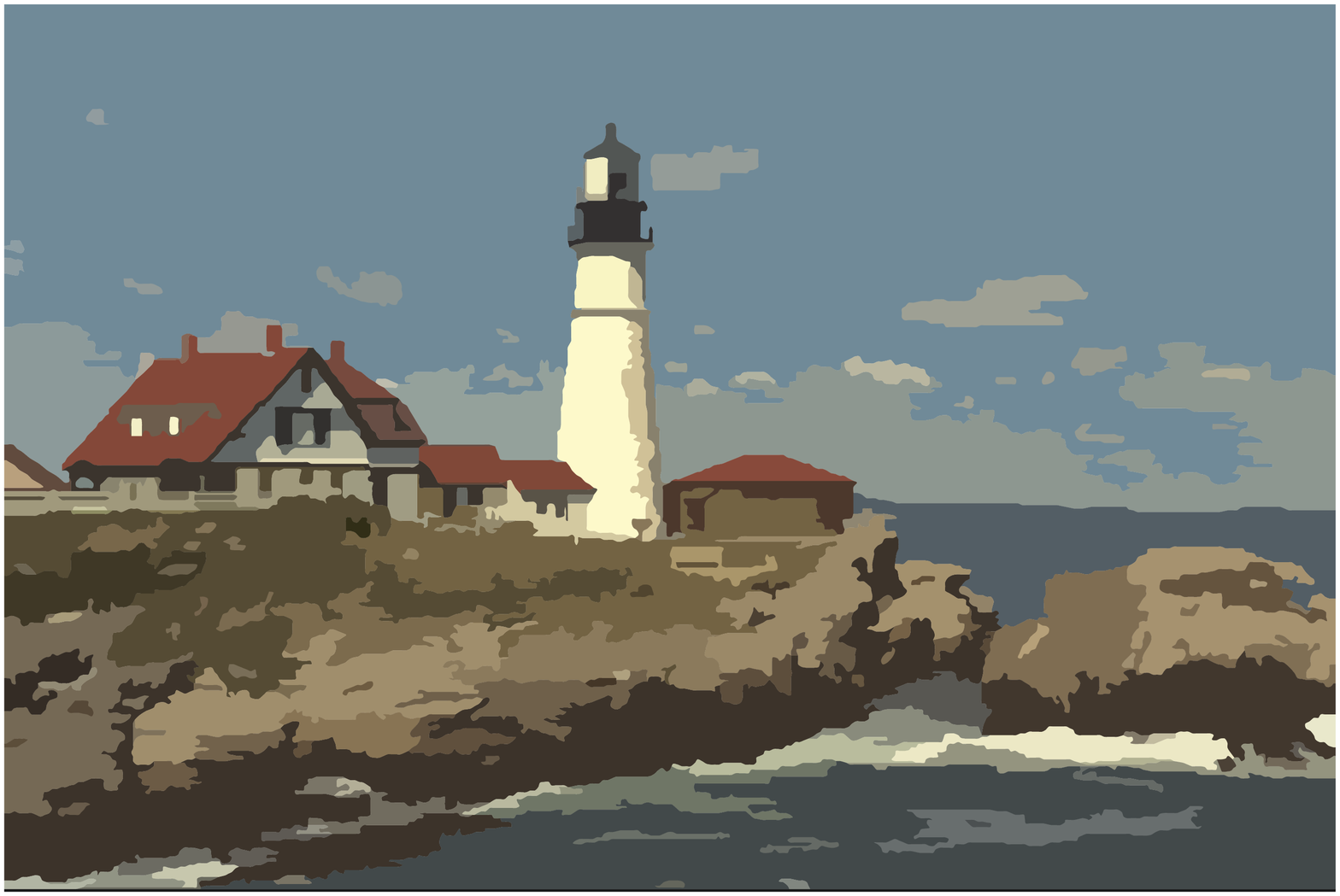}\\
\mbox{a}&\mbox{b}\\
\includegraphics[width=0.4\linewidth]{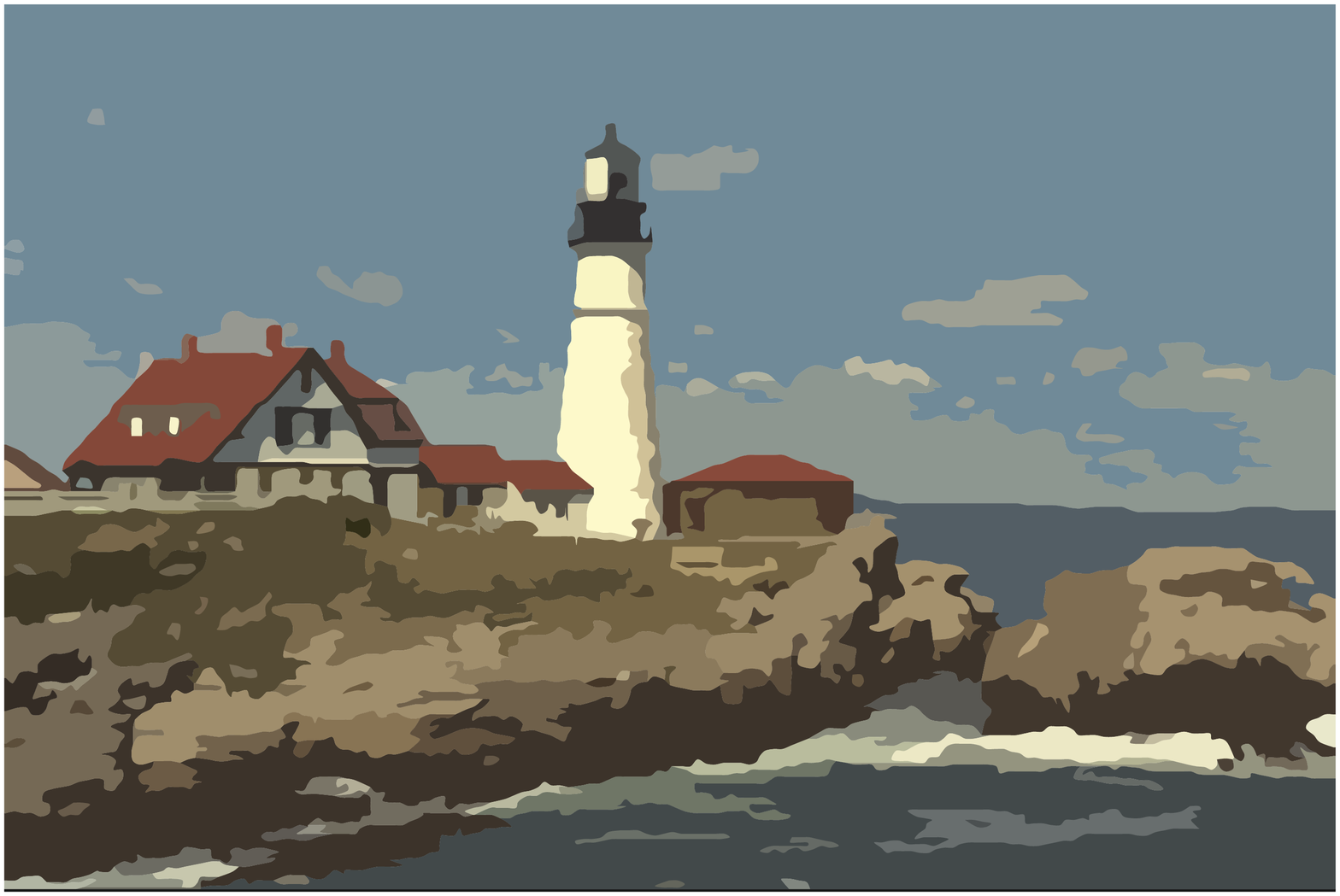}&
\includegraphics[width=0.4\linewidth]{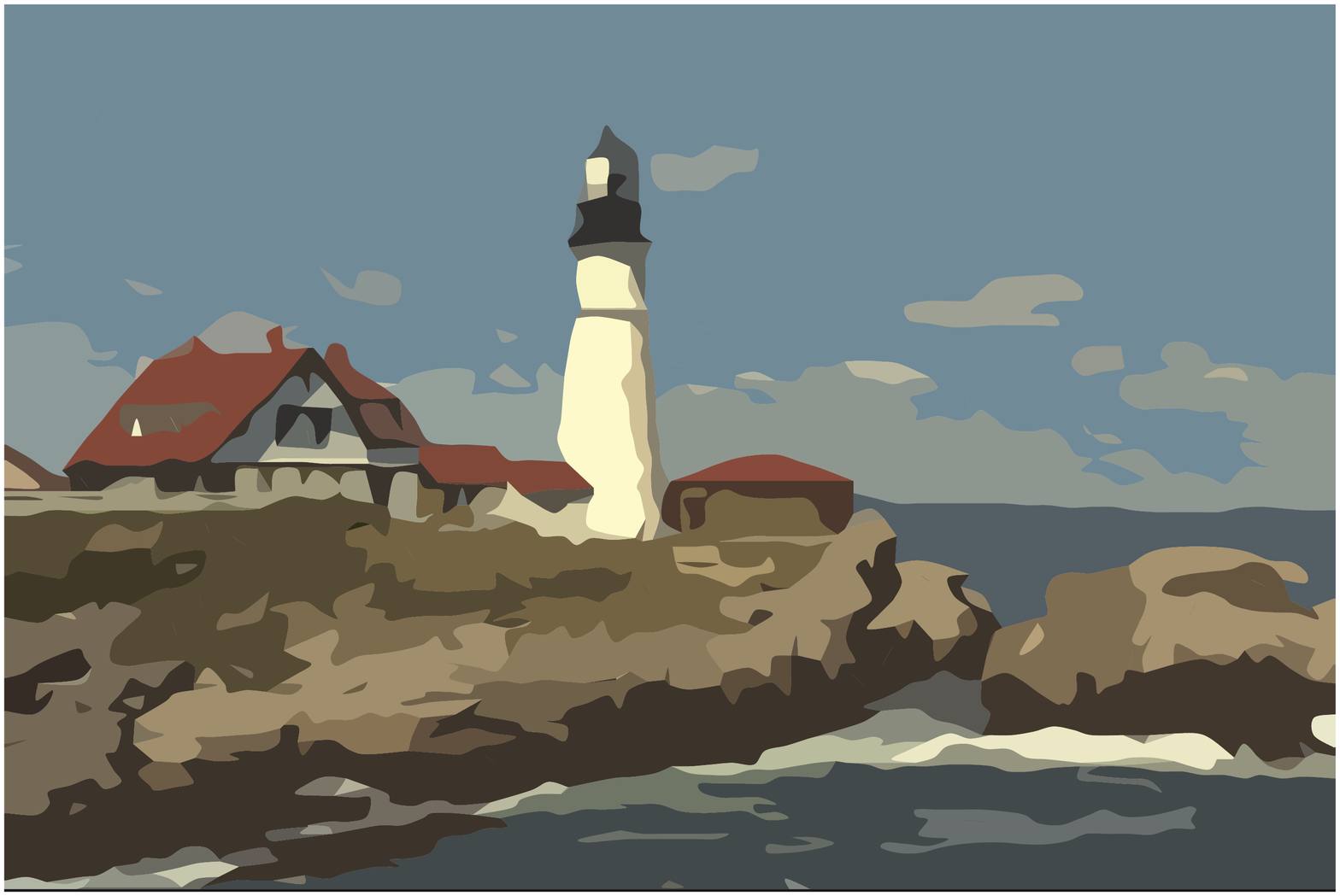}\\
\mbox{c}&\mbox{d}\\
\includegraphics[width=0.4\linewidth]{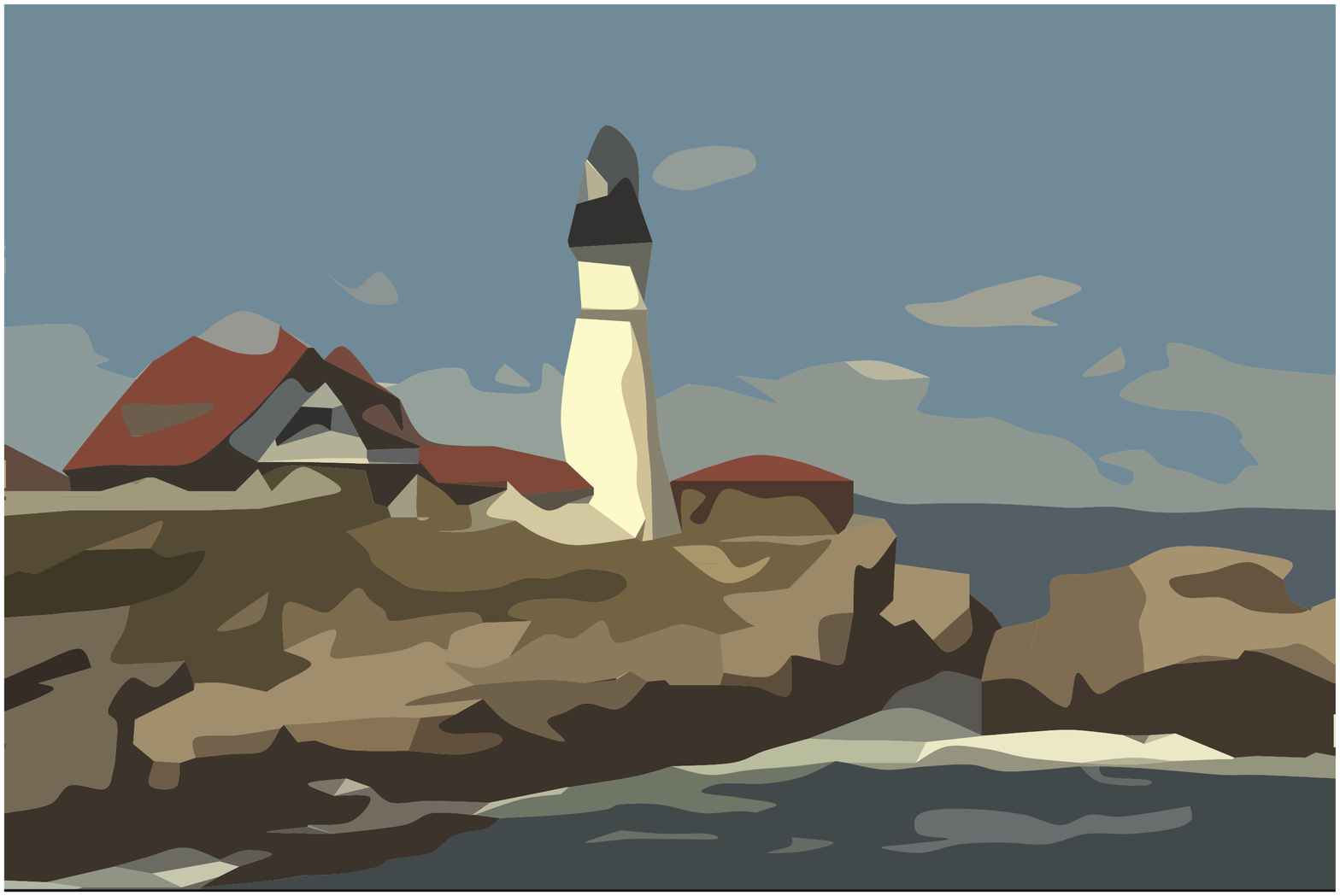} &
\includegraphics[width=0.4\linewidth]{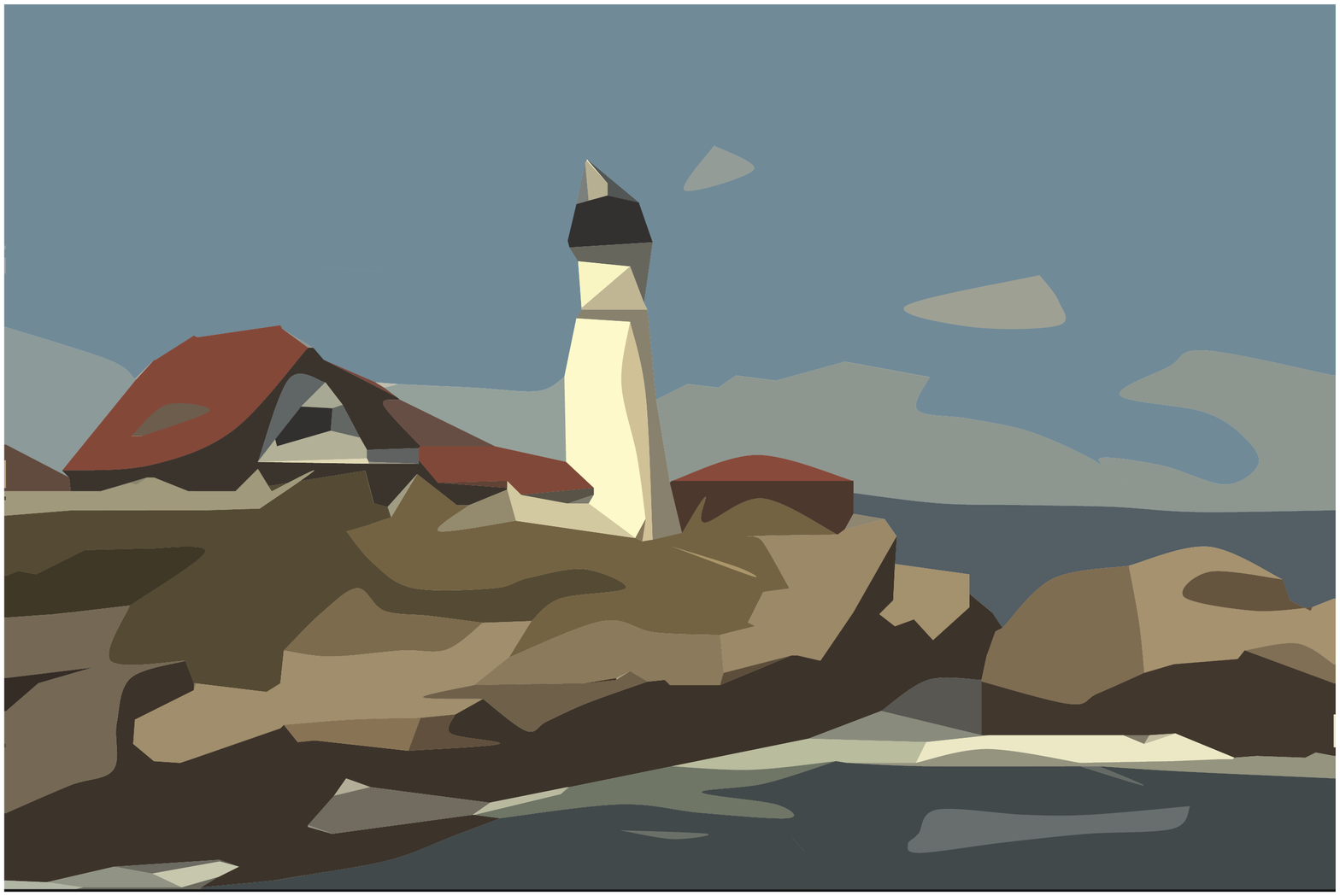}\\
\mbox{e}&\mbox{f}\\
\end{array}$
\caption{Vectorization results of \emph{(a)} with sampling resolution values: \emph{(b) $1/10$, (c) $1/20$, (d) $1/50$, (e) $1/100$, (f) $1/200$}}
\label{fig:six}
\end{center}
\end{figure}

In our method, sampling resolution of control point vector $P$, which is described in section 3.2, determines the smoothness of each fitted curve. This situation can be observed in Figure~\ref{fig:six} for different control point resolutions. As can be seen in (a) and (b), selecting control points close on a curve lets you preserve the shape of a region. However, this is not the desired situation all the time. Since most artists prefer less complex region borders and less detail to obtain an abstract model of the input image, our method has a great advantage over others. By selecting less control point on any curve, one can get rid of unnecessary or unwanted detail as shown in (c) and (d) in Figure~\ref{fig:six}.

\begin{figure}[h]
\begin{center}
$
\begin{array}{cc}
\includegraphics[width=0.4\linewidth]{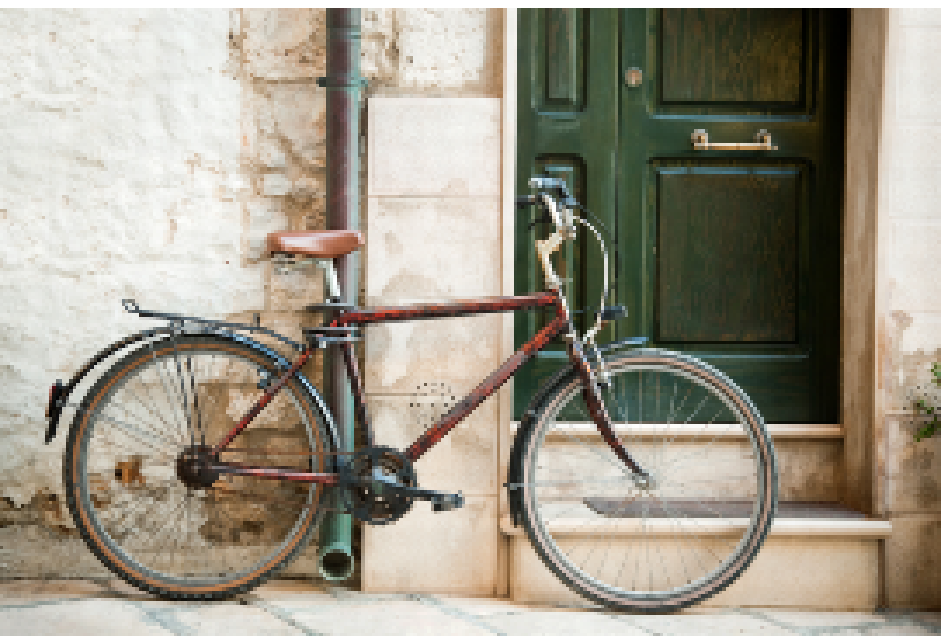}&
\includegraphics[width=0.4\linewidth]{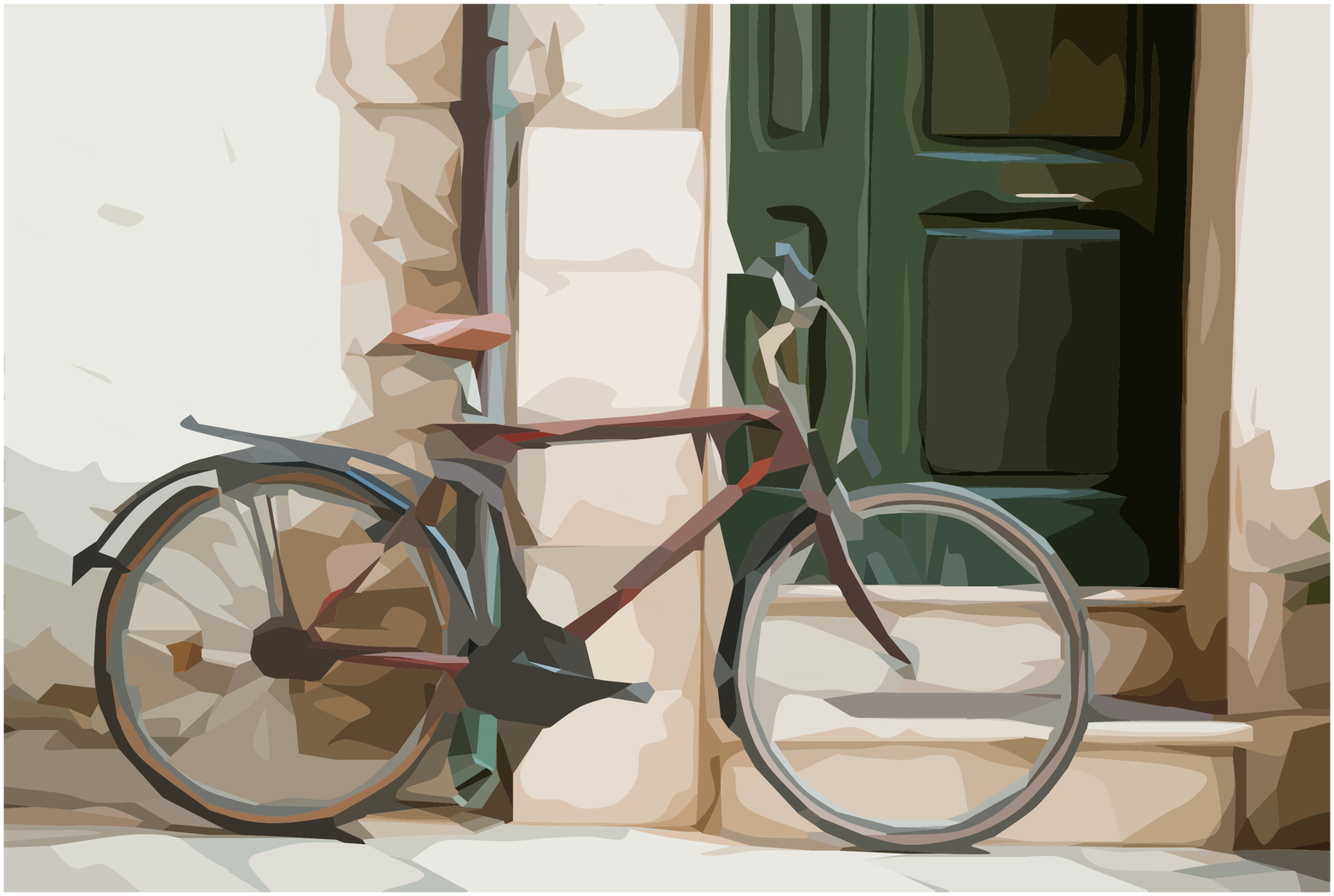}\\
\mbox{a}&\mbox{b}\\
\includegraphics[width=0.4\linewidth]{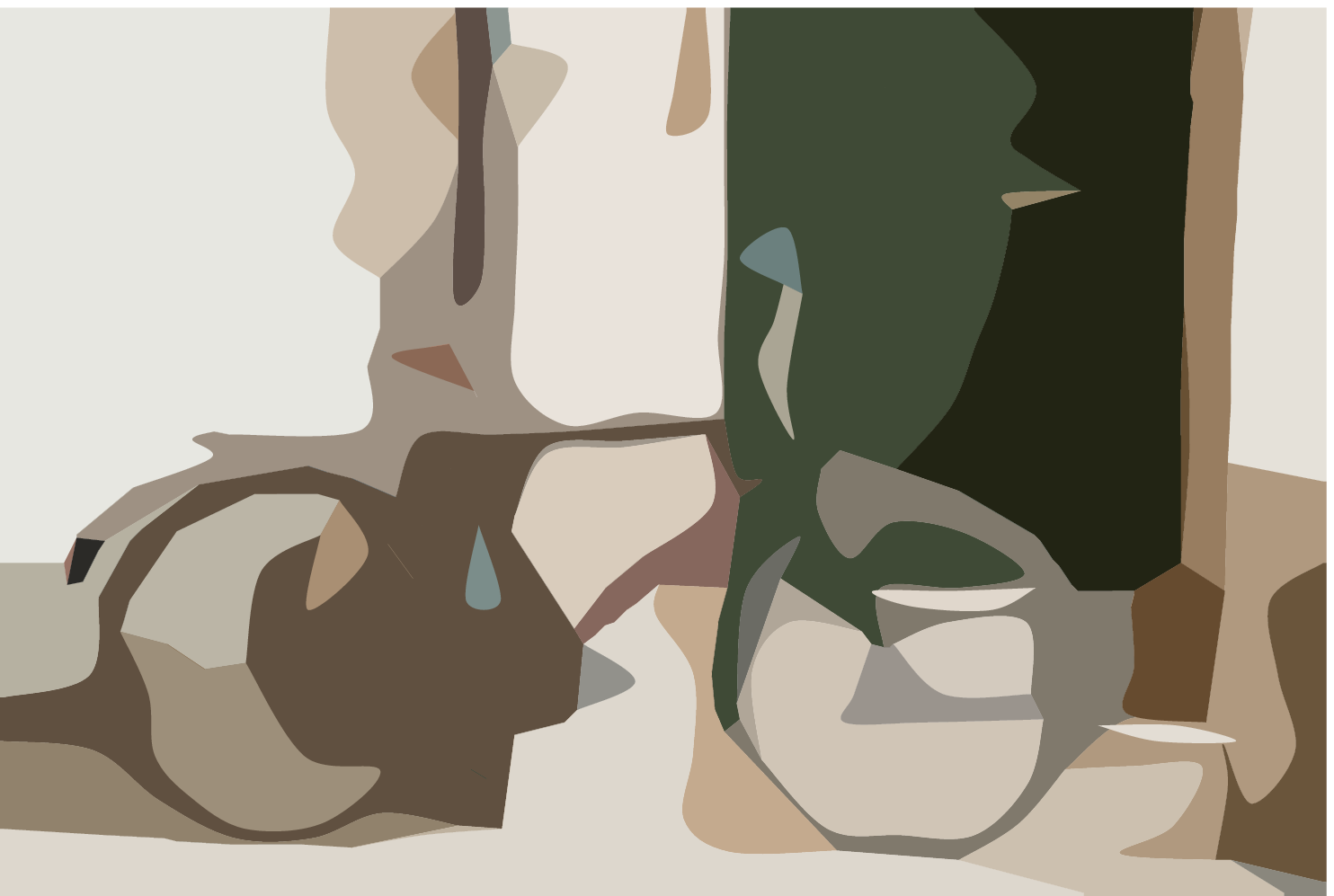}&
\includegraphics[width=0.4\linewidth]{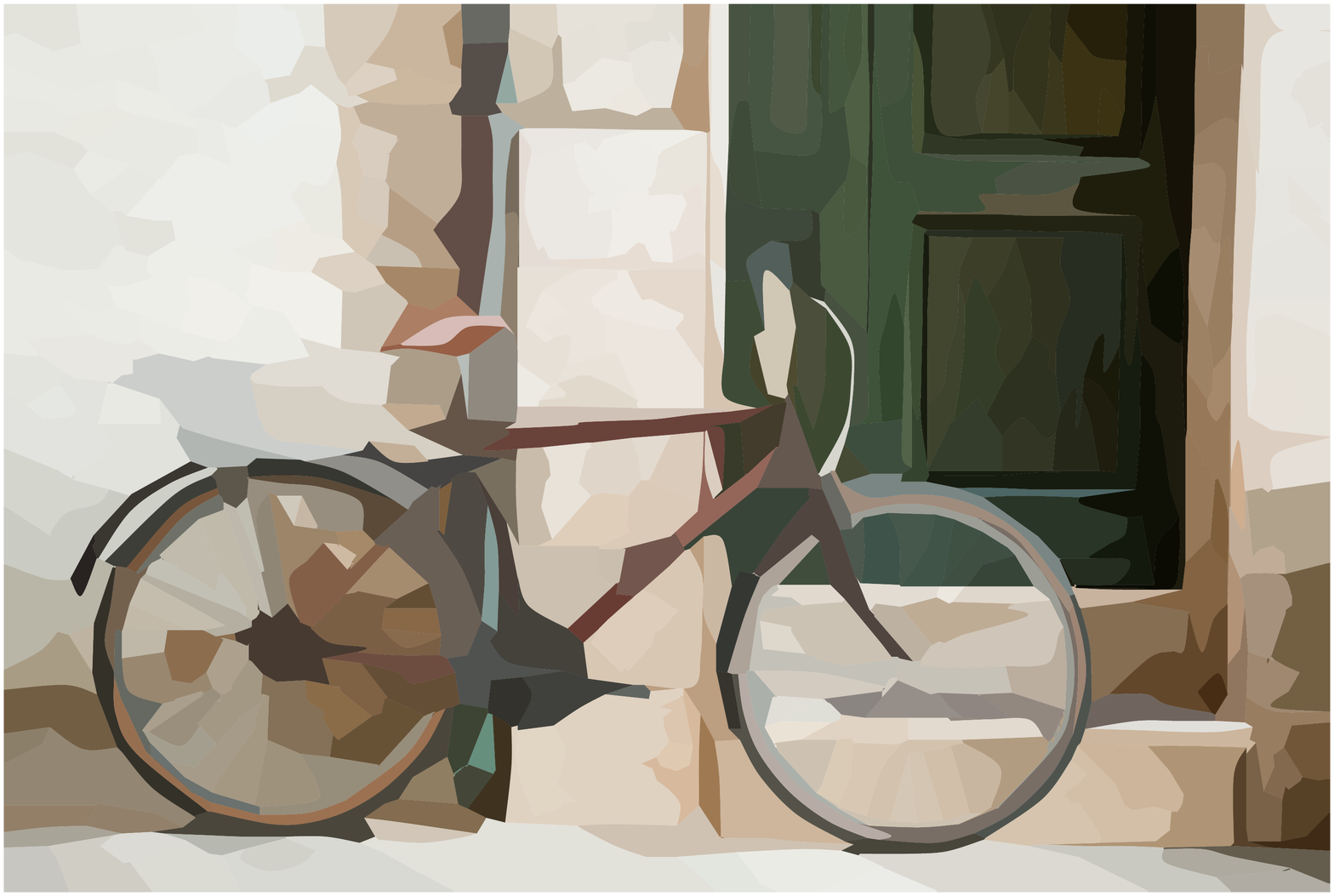}\\
\mbox{c}&\mbox{d}\\
\end{array}$
\caption{(a) Input image, (b) Sample Vectorization using SRM, (c) Sample Vectorization using (CSC), (d) Sample Vectorization using Graphs }
\label{fig:seven}
\end{center}
\end{figure}

Also, the type and amount of segmentation determines the amount of detail. In Figure~\ref{fig:seven}, various vectorization outputs obtained by making use of different segmentation techniques, which are described in section 4, are shown. The parameters of segmentation operations are selected intuitively, whereas the control point selection resolution parameter of vectorization operations are set to the same value ($1/100$) for (b), (c), (d). It can be easily seen that flexibility of using any segmentation technique with our method, gives the freedom to obtain a wide variety of styles.

\begin{figure}[h]
\begin{center}
 \includegraphics[width=1\linewidth]{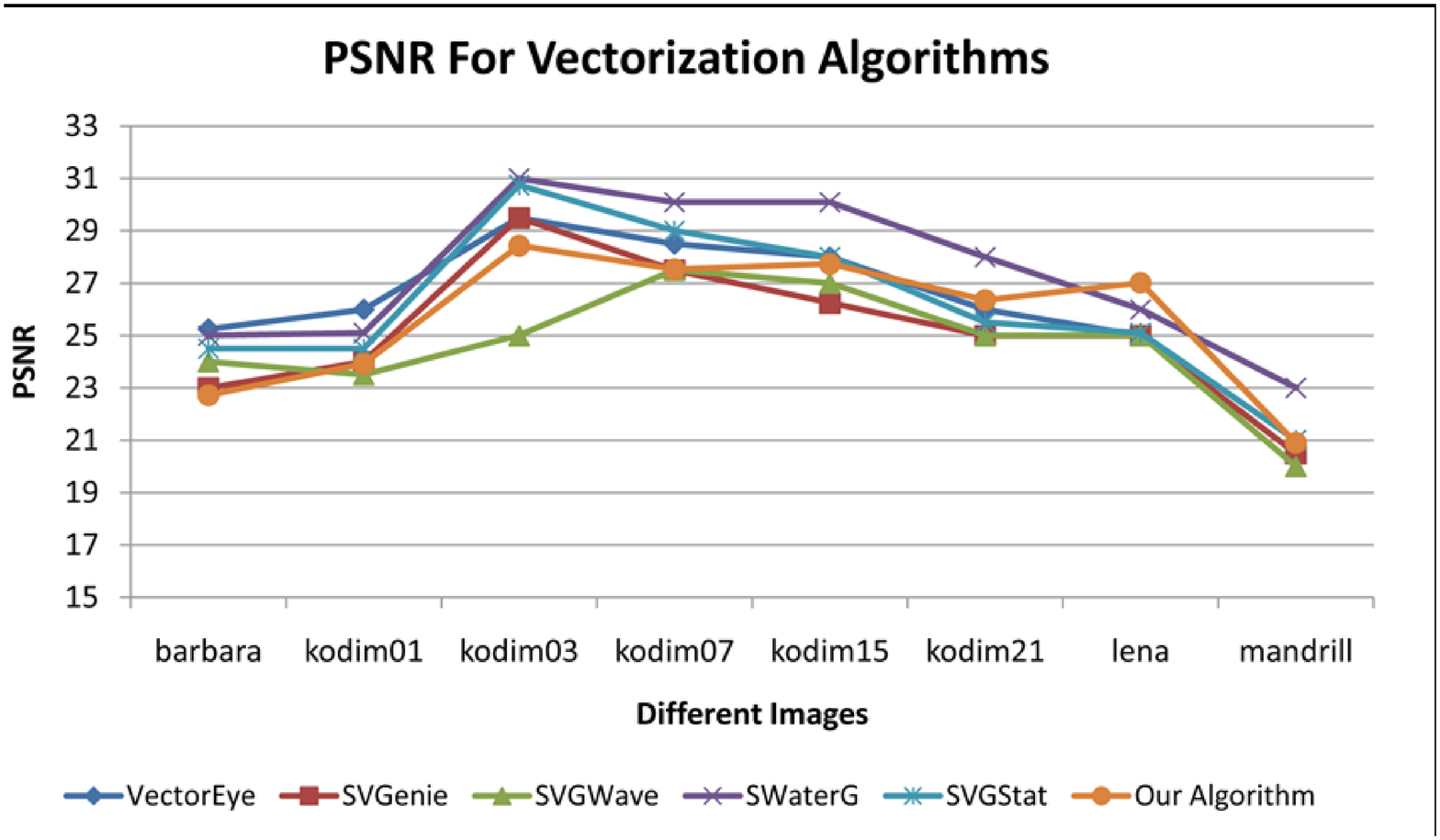}
\end{center}
   \caption{PSNR scores of various vectorization results.}
\label{fig:eight}
\end{figure}

Another evaluation aspect, which is memory efficiency of the way an image is represented, is evaluated and compared with other vectorization methods in literature. To be able to do this, some commonly used test images are vectorized by setting necessary parameters in order to obtain outputs which are visually comparable with the outputs of other methods. Some commonly used test images are determined from \cite{svgstat}, \cite{Watershed}. When Figure~\ref{fig:eight} is examined, it is seen that the vector images are similar Peak Signal-to-Noise Ratio (PSNR) scores.

\begin{figure}[h]
\begin{center}
 \includegraphics[width=1\linewidth]{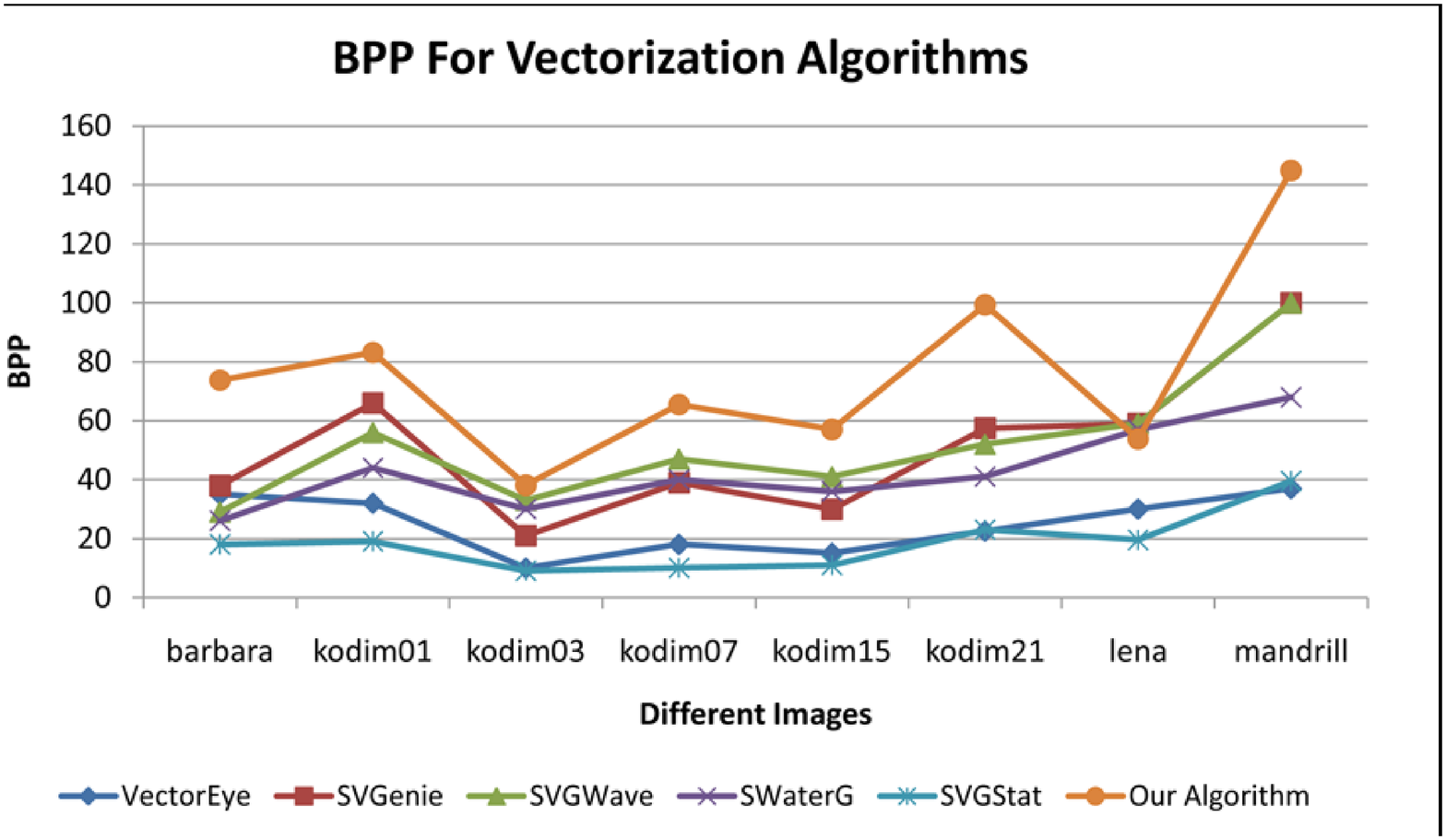}
\end{center}
   \caption{bpp scores of various vectorization results.}
\label{fig:nine}
\end{figure}

At this point, the efficiency of the algorithm can be observed from its bits per pixel (bpp) value. bpp scores of compared vector images are given in Figure~\ref{fig:nine} which is taken from \cite{svgstat} for other vectorization methods. As seen from Figure~\ref{fig:eight} and Figure~\ref{fig:nine}, our method has slightly worse efficiency rank among the vectorization tools compared. This is due to the determined operational priorities which are aesthetics and operation speed. Further studies can be performed to optimize the bpp values, but we will not mention it in this paper. Instead, we will evaluate its run time performance.

\subsection{Performance Evaluation}
There is no significant cost of the algorithm other than the junction point extraction, which is  $O(4 * N^2)$. Since the rest of the algorithm operates directly only on the extracted contours between the junction points, it's efficient both in terms of memory and cpu resources. Such a property makes it suitable for (soft) realtime applications, where speed is an important concern. We observed that vectorization algorithm works realtime in small webcam images with sizes under 480x360 approximately. The algorithm is only sensitive to the number of segmented regions, or in other words to number of paths connecting the junction points. If the image is over-segmented the number of junction points, and hence the number of paths increase. That causes a slow down in the overall runtime. Luckily, the path construction part has a parallel nature. Each spline fit can be performed independently. Hence the algorithm can be parallelized both on CPUs and GPUs. This parallelization overcomes the problem of over-segmentation, and causes the algorithm run reasonably fast even if the number of splines increases. Obviously, so far we have ignored the cost of image segmentation, and rendering the output vector image. One should have a fast segmentation algorithm when speed is important.

\section{Conclusion}
In this paper, we proposed a novel approach for soft real-time vectorization of raster images, using spline approximation of segmentation boundaries. We noted that the algorithm is highly customizable through using different segmentation methods, control point selection techniques and choosing different parameters, such as control point distance. Moreover, we mentioned that our algorithm is reasonably easy to implement, easy to parallelize, robust and generates small svg files when spline approximation is optimized. Finally, it enjoys the smoothness of spline curves on segmentation boundaries, giving the output a unique aesthetic look. Further research will be on selection of control points, using different splines, parallelization with hard real-time constraints. We will also try to combine our technique with other methods in the literature to come up with faster and perceptually more meaningful results.

\bibliographystyle{acmsiggraph}
\nocite{*}
\bibliography{template}

\end{document}